\documentclass{article}
\usepackage{arxiv}

\usepackage{hyperref}
\usepackage{bm}
\usepackage{cite}
\usepackage{graphics,fancyhdr,graphicx,subfigure}
\usepackage{subfigure}
\usepackage{amsthm,amsmath,amsfonts,latexsym,amssymb}
\usepackage{lineno}
\usepackage{diagbox}
\usepackage{algorithm}
\PassOptionsToPackage{noend}{algpseudocode}
\usepackage{algpseudocode}
\usepackage{multirow,booktabs}
\usepackage{listings}
\usepackage[table]{xcolor}
\usepackage{lscape}
\usepackage{comment}
\usepackage{tabularx}
\usepackage{graphicx}
\usepackage{tabularx}
\usepackage{arydshln}
\usepackage{nomencl}
\usepackage{array}
\usepackage{caption}
\usepackage{breqn}
\usepackage{lineno}
\usepackage{mathtools}
\usepackage{subfigure}
\usepackage{caption}
\usepackage{lmodern}
\usepackage{calligra}
\usepackage{xspace}
\usepackage[utf8]{inputenc}
\usepackage{enumerate}
\usepackage[english]{babel}

\def\equationautorefname~#1\null{%
  Eq.~(#1)\null
  }
\def\subfigureautorefname~#1\null{%
  Fig.~#1\null
}

\definecolor{listinggray}{gray}{0.9}
\definecolor{lbcolor}{rgb}{0.9,0.9,0.9}
\definecolor{Darkgreen}{RGB}{0,100,0}

\title{Physics informed WNO \thanks{https://www.csccm.in/}}

\author{ \hspace{1mm}Navaneeth~N.\\
	Department of Applied Mechanics\\
	Indian Institute of Technology (IIT) Delhi\\
	Hauz Khas - 110 016, New Delhi, India \\
	\texttt{navaneeth.n@am.iitd.ac.in} \\
	\And
        \hspace{1mm}Tapas~Tripura\\
	Department of Applied Mechanics\\
	Indian Institute of Technology (IIT) Delhi\\
	Hauz Khas - 110 016, New Delhi, India \\
	\texttt{navaneeth.n@am.iitd.ac.in} \\
	\And
	\hspace{1mm}Souvik~Chakraborty \\
	Department of Applied Mechanics\\
	Indian Institute of Technology (IIT) Delhi\\
	Hauz Khas - 110 016, New Delhi, India \\
	\texttt{souvik@am.iitd.ac.in} \\
}

\renewcommand{\shorttitle}{Physics informed WNO}

\hypersetup{
pdftitle={Physics informed WNO},
pdfsubject={q-bio.NC, q-bio.QM},
pdfauthor={Navaneeth N., Tapas Tripura, Souvik Chakraborty},
pdfkeywords={Stochastic Projection, Physics informed neural network, Operator learning, Wavelet neural operator},
}
\begin{document}
\maketitle

\begin{abstract}
Deep neural operators are recognized as an effective tool for learning solution operators of complex partial differential equations (PDEs). As compared to laborious analytical and computational tools, a single neural operator can predict solutions of PDEs for varying initial or boundary conditions and different inputs. A recently proposed Wavelet Neural Operator (WNO) is one such operator that harnesses the advantage of time-frequency localization of wavelets to capture the manifolds in the spatial domain effectively. While WNO has proven to be a promising method for operator learning, the data-hungry nature of the framework is a major shortcoming. In this work, we propose a physics-informed WNO for learning the solution operators of families of parametric PDEs without labeled training data. The efficacy of the framework is validated and illustrated with four nonlinear spatiotemporal systems relevant to various fields of engineering and science.  
\end{abstract}


\section{Introduction}
\label{sec:intro}
Natural systems are governed by certain conservation and constitutive laws. Modeling of complex constitutive laws using Partial Differential Equations (PDEs) is a prevalent approach across various corners of science and engineering disciplines \cite{debnath2005nonlinear,jones2009differential,evans2010partial}. For a complete understanding of the dynamical behavior of the underlying physical system, it is often required to solve the governing model. In most cases due to the absence of analytical solutions, numerical computational tools such as Finite Element Methods (FEM) \cite{kang1996finite}, isogeometric analysis \cite{cottrell2009isogeometric}, finite difference methods (FDM) \cite{ozicsik2017finite}, and finite volume methods (FVM) \cite{eymard2000finite} are preferred. In addition to solving the system of PDEs, scientific and engineering explorations sometimes involve solving the underlying systems for different input parameters such as different domain geometry, source functions, and initial and boundary conditions (ICs and BCs), often called parametric PDEs. In this scenario, the numerical tools prove to be a computationally expensive alternative since the abovementioned schemes are highly expensive and require independent runs for each combination of input parameters. To enhance the available computational approaches data-driven surrogate approaches such as deep neural networks (DNNs) \cite{sirignano2018dgm,chan2019machine}, and deep energy methods \cite{samaniego2020energy} are proposed. 
The emergence of such data-driven approaches as popular scientific machine learning (Sci-ML)  techniques for learning solutions of PDEs in different areas of science and engineering is evident from the literature \cite{psichogios1992hybrid,lagaris1998artificial,sun2020surrogate,zhu2019physics}. These methods have proven to be computationally superior and on par in terms of accuracy with classical schemes. The concurrent Sci-ML approaches for learning PDE solutions may be broadly categorized into two classes, data-driven and physics-informed. While the first class of approaches aims to learn the solution of a PDE from the data \cite{lu2019deeponet,wu2020data}, the second class of approaches solves the PDE directly from the governing physics, known as Physics Informed Neural Networks (PINN) \cite{raissi2019physics,cai2021physics,chakraborty2021transfer}. The data-driven neural networks (NNs) can learn almost everything \cite{hornik1989multilayer}, however, they require a large amount of labeled training data for effective learning of the solution. Obtaining the paired labeled training data becomes a bottleneck for the data-driven NNs, and these approaches can learn the solution for a fixed combination of input-output only.

To circumvent the need for labeling the training datasets physics-informed learning was later proposed as a viable alternative to the data-driven NNs. The PINN enables the NNs to learn the solution by constraining the physical laws described in the form of differential equations \cite{raissi2019physics,chakraborty2021transfer}. While the network is trained by minimizing the total loss function, comprised of supervised boundary and initial condition loss and a residual loss computed directly from the governing equations, the derivatives involved in the residual loss are seamlessly obtained through the automatic differentiation \cite{margossian2019review}. Later the variational-PINN was proposed, where instead of minimizing the residual of the governing differential equations the variational energy of the system was minimized \cite{goswami2020transfer,samaniego2020energy,kharazmi2021hp}. Despite effectively solving forward and inverse problems \cite{yuan2022pinn}, PINN and variational-PINN have the same inherent limitation as data-driven NNs, i.e., the network can only be trained for fixed input parameters. Once the conditions or the PDE parameters are changed, a retraining of the network is required. Consequently, PINN is impracticable for learning the solutions of parametric PDEs. 

One of the plausible approaches to overcome the aforementioned shortcoming associated with the data-driven and PINN-based NNs is to employ a deep operator-based framework \cite{li2018sequential,kovachki2021neural}. The neural operators are a class of neural networks that learn the solution of the family of parametric PDEs through functional mapping between infinite-dimensional parameter solution spaces. Often the definition of solution maps can be redefined as learning the parameterized kernels of integral transforms from the corpus pairs of the input functions and corresponding solution space. DeepONet was the first neural operator to be introduced as a corollary of the universal approximation theorem for operators \cite{chen1995universal} in which a neural network-based framework is utilized to learn the functional mapping between the input and output paired datasets. Fundamentally, the DeepONet \cite{lu2019deeponet,lu2021learning} is devised of two networks, which enable functional mappings between the input and output, namely branch net and trunk net. While the branch net receives the input function, the trunk net learns the corresponding output at a given sensor point. Another work on graph neural network-based operator learning is presented in \cite{li2020neural}. Here, in contrast to the DeepOnet, Graph Neural Operators (GNO) learns the kernel of integral transform through a message-passing interface between the graph networks. While GNO provides a radically different approach to operator learning, with an increase in hidden layers, the architecture becomes unstable. In more recent work, Li \textit{et al.} have developed another operator learning framework called Fourier Neural Operator (FNO) \cite{li2020fourier,wen2022u}, where the parameters of the integral kernel are learned in Fourier space. Here spectral decomposition enables the framework to obtain the integral kernel in the Fourier space. 
The performance of FNO is promising in comparison with most of the state of art methods of operator learning. However, FNO lacks information about spatial resolution as the basis functions of FFT are frequency localized. This causes poor performance of FNO in the case of PDEs with complex geometry or, in general, in learning the spatial behavior of any signal. 

Motivated by the aforementioned limitation associated with the FNO, Wavelet Neural Operator (WNO) \cite{tripura2023wavelet,thakur2022multi} was proposed where the wavelet transformation \cite{zhang2017time,boggess2015first} was exploited to learn the variation in the input patterns over spatial coordinates through spatial and frequency localization.
Perhaps, WNO can be considered a more generalized case of the previously proposed FNO. The WNO decomposes the input function space into high and low-frequency components through multi-level wavelet decomposition. The outputs at lower levels are contaminated with signal noises, whereas the outputs at higher levels preserve the relevant information. Therefore WNO performs the parameterization of the neural network weights at higher levels of wavelet decomposition only, facilitating accurate and efficient learning of the data. Though WNO is able to learn the highly nonlinear differential operators very effectively regardless of the geometry, like any other data-driven learning method, WNO also requires a sufficiently large data set to effectively learn the solution of an underlying PDE. In most of the practical scenarios, the data set requires corpus pair of input and the corresponding solution, which are often obtained from exorbitantly expensive experiments or time-consuming simulations. In this work, we propose a novel framework that integrates the benefits of WNO with the information from physics to circumvent the drawbacks of the individual frameworks. Evaluating partial derivatives with respect to inputs in kernel-based NNs is a major bottleneck. We overcome this by evaluating the partial derivatives involved in the governing differential equation through stochastic projection-based gradient estimation schemes \cite{navaneeth2023stochastic}. As a result, we obtain a physics-informed wavelet neural operator (PIWNO), which is robust in many senses. For example, the proposed PIWNO learns the underlying solution operator without labeled training datasets (achieving 100\% data efficiency). This is a major disadvantage in all the existing operator learning frameworks. Unlike conventional data-driven and physics-informed NNs, the proposed PIWNO learns a family of parametric PDEs, thereby can accommodate an arbitrary set of input parameters, ICs, and BCs without retraining the network. Backed by the governing physics, the proposed PIWNO also generalizes to unseen input parameters. Some key features of the proposed approach include:
\begin{itemize}
    \item The framework requires no training data and governing physics is incorporated through residual loss.
    \item  A better generalization of WNO is accomplished if the training is done with  
    the physics loss, along with fewer data. 
    \item Derivatives of the neural operator involved in the training loss are efficiently achieved through stochastic projection-based gradients.
\end{itemize}
\par
A trained PIWNO is also three to five times as efficient compared to traditional FEM and FDM-based PDE solvers. We have illustrated the performance of the proposed PIWNO on three distinct examples commonly used in modeling transport and fluid dynamics, wave propagation in neurons, static electricity, and phase-field modeling. With the hands-on features, the proposed PIWNO has the potential to accelerate scientific explorations in the areas of computational physics, computational mechanics, computational biology, and other areas of engineering.

The remainder of the paper is organized as follows. The general problem statement is described in the Section \ref{sec:Problem statement}. The proposed approach is elucidated in the following Section \ref{sec: Methedology}. Subsequently, Numerical examples are presented in  the section\ref{sec: Numerical example}. Finally, the conclusion and final notes are provided in Section \ref{sec:Conclusions}.

\section{Problem statement}\label{sec:Problem statement}
The primary goal of this research is to provide a framework for operator learning using WNO that learns directly from governing physics. Here the physics-informed WNO seeks to learn a family of parametric PDEs rather than a single PDE. We proceed further by considering the function spaces, $\mathcal{A}$ and $\mathcal{U}$, containing all the inputs and outputs $\bm{a} \in \mathcal{A}$ and $\bm{u} \in \mathcal{U}$, respectively. Let there exist a differential operator $\mathcal{N}$ that maps the function spaces to null space, i.e., $\mathcal{N}: \mathcal{A} \times \mapsto \mathcal{O}$, where $\mathcal{O}$ is the null space. Within these function spaces a family of parametric PDEs takes the form:
\begin{equation}\label{operator1}
    \mathcal{N}(\bm{a},\bm{u}) = \bm{0}, \,\, \text { in } D \subset \mathbb{R}^d .
\end{equation}
Here, the PDE is defined on an $d$-dimensional bounded domain, $D \in \mathbb{R}^{d}$, with a boundary $\partial D$, where the boundary condition is expressed as:
\begin{equation}\label{operator_BC}
    \bm{u} = \bm{g}, \,\, \text { in } \partial D .
\end{equation}
The parameter $\bm{a} \in \mathbb{R}^{a}$ denotes the input function space and $\bm{u} \in \mathbb{R}^{u}$ denotes the solution space of the parametric PDE. For a fixed domain $x \in D$, the input function space of the operator contains the source term 
$f(x, t): D \mapsto \mathbb{R}$, the initial condition $u(x, 0): D \mapsto \mathbb{R}$, and the boundary conditions $u(\partial D, t): D \mapsto \mathbb{R}$, while the output function space comprises the solution of the parametric PDE, $u(x, t): D \mapsto \mathbb{R}$, with $t$ being the time coordinate. For the differential operator $\mathcal{N}$ and above input features, there exists an integral operator $\mathcal{D}: \mathcal{A} \mapsto \mathcal{U}$, which maps the input functions to the solution space.
As in the case of data-driven WNO, once we have access to the $N$ number of corpus pairs of observations of input and output $\{\bm{a}_{j},\bm{u}_{j}\}^{N}_{j=1}$, a surrogate modeling approach can be employed to approximate the operator. In the context, formulation yields the nonlinear integral operator, $\mathcal{D}$, approximated by the neural network as: 
\begin{equation}\label{neural operator}
    \mathcal{D}:\mathcal{A}\times \bm{\theta}_{NN} \mapsto \mathcal{U},  
\end{equation}
where $\bm{\theta}_{NN}$  represents the trainable parameter space of the neural network. Here, we note the underlying assumption for the operator learning that for any $\bm{a} \in \mathcal {A}$, there exists a unique solution $\bm{u} = \bm{\theta}_{NN}(\bm{a}) \in \mathcal{U}$. With $N$ samples of input-output paired data, the appropriate loss function for a pure data-driven framework can be formulated as follows:
\begin{equation}
    \mathcal{L}_{\text {data }}\left(\bm{u}, \bm{\theta}_{NN}(\bm{a})\right)=\left\|\bm{u}-\bm{\theta}_{NN}(\bm{a})\right\|_{\mathcal{U}}^2=\int_D\left|u(x_i)-\bm{\theta}_{NN}(a)(x_i)\right|^2\mathrm{~d} x
\end{equation}
The operator loss, evaluated by the averaging error across all possible inputs, and $n_d$ space discretizations, is expressed as:
\begin{equation}
    \mathcal{L}_{\text {data }}\left(\bm{u}, \bm{\theta}_{NN}(\bm{a})\right) = \frac{1}{N} \sum_{j=1}^N \sum_{i=1}^{n_d}\left|u_j(x_i)-\mathcal{D}\left(a_j, \bm{\theta}_{NN} \right)(x_i)\right|^2 
\end{equation}
Thus optimizing the network for the given loss function yields the optimal network parameters:
\begin{equation}
    \bm{\theta}^{*}_{NN} = \underset{\bm{\theta}_{NN}} {\text{argmin}}\; \mathcal{L}_{\text {data }}\left(\bm{u}, \bm{\theta}_{NN}(\bm{a})\right)
\end{equation}
Now, to implement the physics-informed learning, we consider a differential operator $\mathcal{N}$ corresponding to the equation \autoref{operator1}, expressed in the residual form as:
\begin{equation}\label{residual1}
    \mathcal{N} (\bm{a},\bm{u};\bm{\theta}_{NN}) = 0
\end{equation}
where $\mathcal{N}$ contains all the derivatives of space and time, functions of $\bm{x},t$ and $\bm{u}$, and input parameters $\gamma$. Thus the \autoref{residual1} can be rewritten as:
\begin{equation}
\mathcal{N}\left(\mathbf{x}, t, \bm{u}, \partial_{\mathbf{t}} 
\bm{u}, \partial_{\mathbf{t}}^2 \bm{u} \ldots, \partial_x \bm{u}, \partial_t^n \bm{u}, \ldots, \partial_x^n \bm{u}, \boldsymbol{\gamma}\right)=0
\end{equation}
%
Here the constraint in \autoref{residual1} is evaluated using the PDE loss along with boundary loss and initial condition loss. 
%
Employing a mean squared loss function, the expression of $\mathcal{L}_{\text{pde}}$ follows the form
\begin{equation}\label{Lpde}
\mathcal{L}_{\text {pde }}\left(\mathcal{D}(\bm{a},\bm{\theta}_{NN})\right) = \underbrace{ \left\|\mathcal{N}(\bm{a},\bm{u};\bm{\theta}_{NN})\right\|_{D}^2}_{\text{Physics Loss}} + \alpha \underbrace{ \left\|{\bm{u}(\bm{x};\bm{\theta}_{NN}) \mid}_{\partial D}-\bm{g}\right\|^2}_{\text{Boundary Loss}} 
\end{equation}
For $N$ training input samples, with $n_d$ space discretizations and $n_b$ boundary points, the \autoref{Lpde} can be rewritten as 
\begin{equation}
\mathcal{L}_{\text {pde }}\left(\mathcal{D}(\bm{a},\bm{\theta}_{NN})\right) =
\frac{1}{N} \sum_{j=1}^N \sum_{i=1}^{n_d}\left|\mathcal{N}(a_j,u_j;\bm{\theta}_{NN})(x_i)\right|^2 + \alpha \sum_{j=1}^N \sum_{i=1}^{n_b}\left|u_j(x_i;\bm{\theta}_{NN})-g(x_i)\right|^2 
\end{equation}
As was mentioned, the residual loss utilizes the stochastic projection-based gradients to obtain the derivatives of the output $\bm{u}$. The derivative of $\bm{u}$ 
 at a given point $\bm {\bar{x}}$ is computed by
\begin{equation}\label{spgf_net}
    \frac{\partial \bm{u}({\bm {\bar x}},\bm{\theta}_{NN})}{\partial {\bm {x}}}= \frac{\frac{1}{N_b}{\sum}_{i=1}^{N_{b}}{( \bm{u}(\bm {x}_i,\bm{\theta}_{NN})- \bm{u}({\bm \bar{x}},\bm{\theta}_{NN}))(\bm {x}_i-\bm {\bar {x}})^{T}}}{\frac{1}{N_b}{\sum}_{i=1}^{N_{b}}{(\bm {x}_i-\bm {\bar{x}})(\bm {x}_i-\bm {\bar{x}})^{T}}}
\end{equation}
where $\bm{x}_i= \{x_i,y_i\}$ is considered to be a generic neighborhood point and $N_b$ represents the number of neighborhood points. A comprehensive description of stochastic projection-based gradients and the implications of the method are provided in the supplementary material.

\section{Proposed Methodology}\label{sec: Methedology}
While the wavelet neural operator learns the solution of the family of parametric PDE,
physics-informed learning aims to learn the operator from physics constraints.
Moreover, since the operator is trained with loss functions computed directly from the governing equations $L_{PDE}$, one needs to evaluate  the derivatives operator output involved in the governing physics. However, in practice, it is non-trivial to obtain the derivatives of the output field,  especially when the architecture of  $\theta_NN$ is devised of the convolution layers. Thus, we conveniently employ a stochastic projection-based method to compute the gradients involved in the training loss. In this section, we provide a high-level description of the WNO and stochastic projection-based gradient computation.  
\subsection{Wavelet Neural Operator}
The WNO is a data-driven neural operator for learning the integral operator of a family of parametric PDEs. In parametric PDEs, we denote the input and output function spaces by $\mathcal{A}$ and $\mathcal{U}$. The input function space $\mathcal{A}$, in general, contains different PDE parameters, geometries, and initial and boundary conditions, while the output space $\mathcal{U}$ denotes the solution space.
Mathematically, we consider a smooth $d$-dimensional domain $D$ over which the functional spaces are constructed. We define the input-output pair $\{ a(x) \in \mathcal{A}$, $u(x) \in \mathcal{U}\}$ such that $u(x) = \mathcal{D}(a(x))$.
Over the domain, we take a relatively simple example of the form,
\begin{equation}
    \begin{aligned}
    -\partial_{xx} u(x)+\omega u(x) & = f\left(x, u(x)\right); \quad x \in D, \\
    u(x) & =0; \quad x \in \partial D,
    \end{aligned}
\end{equation}
where $\omega$ is the parameter of the PDE and $f:D \times \mathbb{C} \rightarrow \mathbb{C}$ is a continuous function. For a valid value of $\omega$ the above differential equation may be recast as an integral equation of the following form,
\begin{equation}
    u(x)=\int_D k(x,\xi) f\left(\xi, u(\xi)\right) d\xi; \quad x \in D,
\end{equation}
where $k(\cdot) \in \mathbb{C}$ is continuous and non-negative Green's function. For nonlinear parametric PDEs, the above integral equation may be generalized to Urysohn-type integral equations,
\begin{equation}\label{eq:urysohn}
    u(x) = \int_{D} k(x, \xi)f\left(\xi, u(\xi)\right) d\xi + g(x); \quad x \in D,
\end{equation}
where $k(\cdot)$ is the kernel of the nonlinear integral equation representing the nonlinear counterpart of Green's function, and $g$ is some linear transformation.
Our aim is to learn the nonlinear integral operator $\mathcal{D}: a(x) \mapsto u(x)$, which maps the input function space $a(x) \in {d_a}$ to the solution space $u(x) \in {d_u}$. From \eqref{eq:urysohn} the input space $a(x)$ is a tuple of the spatial coordinate $x$, the initial conditions $u(x)$, and the forcing function $f(x,u(x))$. 
The data-driven WNO requires a collection of $N$ pairs (number of training samples) of the input-output $\{a_j, u_j\}_{j=1}^N$. In neural network setup, the approximated integral operator $\mathcal{D}$ is given as,
\begin{equation}
    \mathcal{D} : \mathcal{A} \times {\bm{\theta}}_{NN} \mapsto \mathcal{U}.
\end{equation}
Since the integral in \eqref{eq:urysohn} can not be defined in an infinite dimensional space, therefore, a finite-dimensional parameterization space is obtained by discretization of the solution domain $D \in \mathbb{R}^{n}$. 
For featuring a multi-dimensional kernel convolution, the input $a(x)$ is lifted to some high dimensional space ${d_v}$ using a local transformation ${\rm{P}}: a(x) \mapsto v_{0}(x)$. The local transformation ${\rm{P}}$ can be modeled as a shallow fully connected neural network (FNN) or as a $1\times1$ convolution. On the lifted space, $l$-number of iterations of the form in \eqref{eq:urysohn} is performed. The iterations are featured by the transformation $G:\mathbb{R}^{d_v} \mapsto \mathbb{R}^{d_v}$ such that $v_{j+1} = G(v_{j})$. These iterations resemble the usual concepts of hidden layers in neural networks; however, the hidden layers are defined to follow the operator theory in functional analysis \cite{hutson2005applications}. At the end of the $l$-iterations, a second local transformation ${\rm{Q}}: v_{l}(x) \mapsto u(x)$ is applied to obtain the final solution space $u(x) \in \mathbb{R}^{d_u}$. Motivated by the \eqref{eq:urysohn}, the step-wise updates $G(\cdot)$ is defined as follows, 
\begin{equation}\label{eq:iteration}
    G(v_{j})(x):= \varphi \left( \left(K(a; \phi) * v_{j}\right)(x) + W v_{j}(x) \right); \quad x \in D, \quad j \in [1,l],
\end{equation}
where $\varphi(\cdot) \in \mathbb{R}$ is a non-linear activation function, $\phi \in \theta_{NN}$ is the kernel parameters, $W: \mathbb{R}^{d_{v}} \to \mathbb{R}^{d_{v}}$ is a linear transformation, and $K$ is the nonlinear integral operator. Here the integral operator $K$ is defined as,
\begin{equation}\label{eq:integral}
    \left(K(a ; \phi) * v_{j}\right)(x) := \int_{D} k \left(a(x), x, \xi; \phi \right) v_{j}(\xi) \mathrm{d}\xi; \quad x \in D, \quad j \in [1,l].
\end{equation}
where $k \left(a(x), x, \xi; \phi \right)$ denotes the kernel of the nonlinear integral equation in \eqref{eq:urysohn}. The aim is to learn the kernel $k \left(a(x), x, \xi; \phi \right)$ by parameterizing the neural network in the wavelet space. To create the parameterization space in the wavelet domain, wavelet transform is performed on the lifted input $v_{j}(x)$. The forward and inverse wavelet transforms $\mathcal{W}(\cdot)$ and $\mathcal{W}^{-1}(\cdot)$ are defined as follows \cite{daubechies1992ten},
\begin{equation}\label{eq:wavelet}
    \begin{aligned}
        (\mathcal{W} v_{j})(s, \tau) & = \int_{D} \Gamma (x) \frac{1}{|s|^{1 / 2}} \psi\left(\frac{x-\tau}{s}\right) dx, \\
        (\mathcal{W}^{-1} (v_{j})_w)(x) & = \frac{1}{C_{\psi}} \int_{0}^{\infty} \int_{D} (v_{j})_{w}(s, \tau) \frac{1}{|s|^{1 / 2}} \tilde{\psi}\left(\frac{x-\tau}{s}\right) d\tau \frac{ds}{s^{2}},
    \end{aligned}
\end{equation}
where $\psi(x)$ denotes the orthonormal mother wavelet, and $s$ and $\tau$ are the scaling and translational parameters of wavelet decomposition, $(v_{j})_{w}$ is the wavelet decomposed coefficients of $v_{j}(x)$, $\psi(\cdot)$ is the scaled and shifted mother wavelet, and $0 < C_{\psi} < \infty$ is the admissible constant \cite{daubechies1992ten}. Since the aim is to learn the kernel integration in the wavelet domain, the kernel $k_{\phi}$ is directly defined in the wavelet space, denoted as $R_{\phi} = \mathcal{W}(k_{\phi})$. Using the convolution theorem, the integral in \eqref{eq:integral} over the wavelet domain is expressed as
\begin{equation}\label{eq:conv_final}
    \begin{aligned}
        \left(K(\phi) * v_{j}\right)(x)=\mathcal{W}^{-1}\left(R_{\phi} \cdot \mathcal{W}( v_{j})\right)(x); && x \in D.
    \end{aligned}
\end{equation}
During training the network, performing the complex wavelet decomposition in \eqref{eq:wavelet} is expensive since the scale and translation parameters $s$ and $\tau$ can be infinite-dimensional. In order to perform the complex wavelet transform quickly, we utilized the slim dual-tree complex wavelet transform (DTCWT) toolbox in \cite{cotter2020uses}, originally proposed in \cite{selesnick2005dual}. At each decomposition level, the slim DTCWT provides two sets (real and imaginary) of six wavelet coefficients, which roughly represent coefficients of 15$^{\circ}$, 45$^{\circ}$, 75$^{\circ}$, 105$^{\circ}$, 135$^{\circ}$, and 165$^{\circ}$ wavelets. Due to conjugate symmetry, the coefficients are halved at each decomposition level.
To learn the relevant features of the input, the parameterization of the kernel $R_{\phi}(\cdot)$ is done at the highest level of the DTCWT. Thus, if ${d_l}$ denotes the dimension of each wavelet coefficient at the last level of DTCWT, then for the input $v_{j}(x)$, the decomposed output $\mathcal{W}\left(v_{j}; \ell \right)$ will have a dimension ${{d_l} \times d_{v}}$. Thereafter the weight tensor $R_{\phi}(\ell)$ is constructed with the size ${d_{l} \times d_{v} \times d_{v}}$. The kernel convolution $(R_{\phi} \cdot \mathcal{W}( v_{j}))(x)$ follows,
\begin{equation}\label{eq:convolution_final}
    \begin{aligned}
        \left(R \cdot \mathcal{W} (v_{j}; \ell)\right)_{t_1,t_2} = \sum_{t_{3}=1}^{d_{v}} R_{t_1, t_2, t_3} \mathcal{W} (v_{j}; \ell)_{t_1, t_3}; && I_1 \in [1, {d_l}], && I_{2}, I_{3} \in d_{v}.
    \end{aligned}
\end{equation}
where the integration is performed in the uplifting dimension. In total, the DTCWT provides 12 wavelet coefficients, real and imaginary coefficients of 15$^{\circ}$, 45$^{\circ}$, 75$^{\circ}$, 105$^{\circ}$, 135$^{\circ}$, and 165$^{\circ}$ wavelets. There, to learn the parametric space, we need twelve weight tensors and twelve convolutions defined in \eqref{eq:convolution_final}.

\begin{figure}[!ht]
    \centering
    \includegraphics[width=\textwidth]{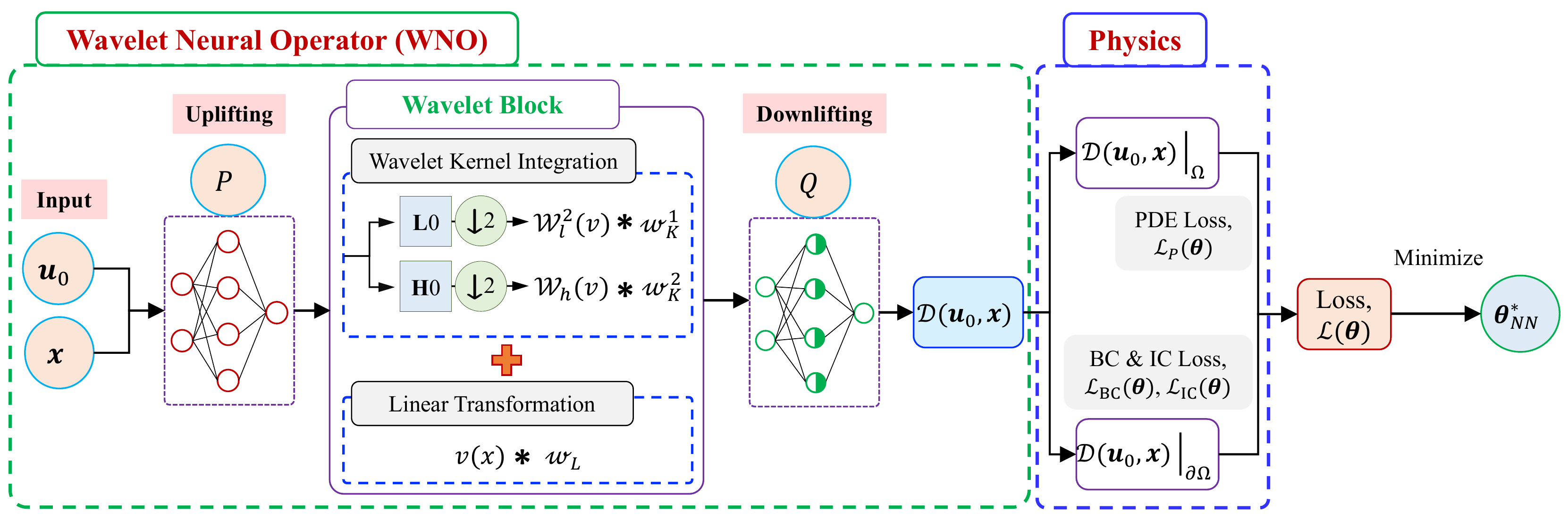}
    \caption{\textbf{Physics-informed Wavelet Neural Operator (PIWNO)}. Motivated by the operator theory in classical functional analysis the WNO architecture aims to learn Green's function of underlying PDEs. In WNO the inputs are first lifted to a high-dimensional latent space, over which certain iterations followed by nonlinear activation are applied. The iterations are represented using wavelet kernel integration blocks. It consists of a kernel integration network that learns the kernel of integration and a linear transformation network that learns the kernel integration constants. For learning the kernel, the latent inputs are localized in space-frequency using wavelets, which is a feature of WNO. The integration outputs and the integral constants are added and finally down-lifted to obtain the solution of underlying PDE. The solutions are constrained to satisfy given PDEs, BC, and IC. For devising the PDE constraint, the spatial derivatives are computed through a stochastic projections-based gradient estimation scheme.}
    \label{fig:piwno}
\end{figure}

The WNO is defined by the following parameters (i) uplifting dimensions, (ii) choice of vanishing moments in the wavelets, (iii) decomposition level, and (iv) the number of wavelet blocks. The uplifting dimension denotes the number of kernels. As in convolution neural networks (CNN), a higher uplifting dimension makes it easy for the network to learn, but too high may lead to overfitting. The vanishing moments represent the smoothness of a wavelet. Ideally, for an input having high spatial variations, wavelets with low-order (in WNO, $<4$) vanishing moments help in capturing the local features. For a relatively smooth image, a wavelet with higher ($>4$) vanishing should be utilized. The number of wavelet decomposition levels $\ell$ depends on the kernel size. If fine features are to be captured, then a smaller kernel is required, meaning a higher decomposition level. It is to be noted that each decomposition subsamples the input size by a factor of two. For e.g., if the input has a spatial dimension $64\times64$ and if we want a kernel of sizes 4 and 8, then we need a decomposition level of 4 and 3, respectively. The number of wavelet blocks depends on the complexity of the underlying operator to be learned. A schematic of the architecture employed for physics-informed WNO is provided in the \autoref{fig:piwno}.

\subsection{Derivatives through stochastic projection}
As we have discussed earlier, the primary requirement to compute $L_{pde}$, which is utilized to train the network, is the evaluation of derivatives of output $\bm{u}$. However, it is not straightforward to compute the derivatives for neural operators efficiently. While there are different methods available to compute the gradients, here we employ stochastic projection-based gradients to obtain the gradients seamlessly. The efficacy of the method is elucidated in \cite{navaneeth2023stochastic}. In principle, the stochastic projection method can be understood as the upscaling of microscopic information in a multi-scale formulation where the micro-scale information accounts for non-trivial directional information of the evolving state variables. Now, to put the stochastic projection in perceptive of physics-informed operator learning, we consider a domain with a finite number of collocation points over which the solution is sought. Also, there is a neighborhood defined by the distance for each collocation point. Now we assume that the variation of the field variable is distinctively measured only up to a certain characteristic distance while the variation of the field variable less than the characteristic length can be  treated to be a stochastic process. While in the former case, the field variable is considered to be a macroscopic field variable, the latter one is considered to be the microscopic field variable. Similarly, the field variables can be differentiated into those which evolve slowly with time and those which evolve on faster time scales based on the characteristic time interval. 

To begin with the formulation, we suppose $\bm u(\bm x)$ to be a macroscopic field variable prior to conditioning based on any microscopically inspired information. Further, to characterize the macroscopic field variable u(x) by means of neighborhood information, a zero mean noise term is added to the $\bm u(\bm x)$, which results in field measurement, $\bm u(\bm z)$ at $\bm z$ such that $\bm z \neq \bm x$, and is expressed as:
\begin{equation}
    \bm u(\bm z) = \bm u(\bm x)+\mathbf \Delta \eta
\end{equation}
Here, the term $\mathbf \Delta \eta$ represents noise due to the unaccounted fluctuations in the microscopic level. The noisy observation at the macroscopic level, after constraining the information from the micro scale,  sampled at a time t, can be written in the following form;
\begin{equation}\label{sampled observation}
    d{\bm {Z}_t} =\bm{h}(\bm{x}_t,\bm{z}_t)d{t}+\sigma{d{\bm{W}_t}}.
\end{equation}
Here we note that, in the above expression, $\bm {h}(\cdot,\cdot)$ is a function that yields the difference in the field variable values at points separated by a distance observed in macroscopic level. While the term $\sigma{d{\bm{W}_t}}$ indicates noise that is reliant on its microscopic counterpart and microscopically sampled function $\bm Z_t$, $\bm W_t$ is Brownian motion that is independent of $\bm eta_t$. To obtain the microscopically informed spatial variation in the field variable, a conditional expectation of $\bm u$ is utilized. Consequently, applying the analogy of stochastic filtering on  microscopically sampled function $\bm Z_{t}$ for a process measurable in a given probability space and further simplification utilizing  the Kallianpur-Stribel formula and Radon-Nikodym  derivative used in the change of measure, the $(\bm{z}_t-\bm{x}_t)$  yields the following form:
\begin{equation}\label{eq11}
    (\bm{z}_t-\bm{x}_t)= (\bm{z}_{t_0}-\bm{x}_{t_0})+  \int_{\hat{t_0}}^{\hat{t}} \left(\pi_{s}\left((\bm{z}-\bm{x}) \bm{h}^{T}\right)-\pi_{s}(\bm{h})^{T} \pi_{t}(\bm{z}-\bm{x})\right) \cdot\left(\sigma \sigma^{T}\right)^{-1}\left(d \bm{z}_{t}\right)
\end{equation}
where $\bm z_{t_0}-\bm x_{t_0}=\Delta$ with $t_0$ being the initial time. The term $\pi_{t} (\bm{z}-\bm{x})$ represents conditional expectation with respect to given probability space $\mathcal P$ and which is expressed by $\Pi_t(\bm u) = E_{\mathcal P}[\bm u(\bm x)|\mathcal F_t]$ such that $F_t$ denotes the sequence formed by adding up $Z_t$ at each time till t. In addition, \autoref{eq11} also satisfies the presumption that the macroscopic field variable cannot be resolved into a vector having length less than the characteristic length, $|\Delta|$. Here, the function, $\bm h(\bm x, \bm z)$ is smooth enough and satisfies
\begin{equation}
    \bm h(x,z) =
    \begin{cases}
      0, & \text{if}\ |z-x|\leq 0 \\
      Nonzero, & \text{otherwise}.
    \end{cases}
\end{equation}
It is important to notice that $\bm Z_t$ in \autoref{eq11} denotes the specific observation: 
$\bm{Z}_t = {\bm{Z}_{t'}} + 
{\int _{\delta \hat{t}}{\Delta}{d\hat{s}}} $ where as $t^{'}$ is the previously sampled macroscopic time.

From the \autoref{sampled observation}, variance of the microscopic observation can be substituted as: $\bm{\sigma} \bm{\sigma}^T \equiv \pi_t\left(\bm{h} \bm{h}^T\right) \delta \hat{t}$. 

Since the integration is carried out within the least microscopically resolvable time period $t$, macroscopic level temporal variations in the integrand are not solvable. Thus, $\bm z_t - \bm x_t$ is considered to be drift-less and can be approximated as:
\begin{equation}
    (\bm{z}_t-\bm{x}_t)\sim \Delta + \mathbf{G}\Delta
\end{equation}
where the expression of $\mathbf{G}$ is given by 
\begin{equation}
    \mathbf G = \left(\pi_{t}\left((\bm{z}-\bm{x}) \bm{h}^{T}\right)- \pi_{t}(\bm{z}-\bm{x})\right)\pi_{t}(\bm h)^{T})((Var(\bm h))^{-1}.
\end{equation}
Now to compute the gradients at a given point ${\bm{\bar{x}}} = \{x_p,y_p\}$ in the domain, a neighborhood is specified in terms of the radius $r_n$. Once the neighborhood is defined, one may choose $N_t$ number of collocation points inside the neighborhood. Subsequently, gradient of output $\mathcal{U}$ with the input variable at $\bm {\bar{x}}$ is computed by \autoref{spgf_net}, where $\bm{x}_i= \{x_i,y_i\}$ is considered to be a generic neighborhood point. 
\begin{equation}\label{spgf_net2}
    \mathbf{\hat G}(\bm x={\bm{\bar{x}}}) = \frac{\partial \bm{u}({\bm {\bar x}}, \bm w)}{\partial {\bm {x}}}= \frac{\frac{1}{N_t}{\sum}_{i=1}^{N_{b}}{( \bm{u}(\bm {x}_i, \bm w)- \bm{u}({\bm \bar{x}}, \bm w))(\bm {x}_i-\bm {\bar {x}})^{T}}}{\frac{1}{N_t}{\sum}_{i=1}^{N_{b}}{(\bm {x}_i-\bm {\bar{x}})(\bm {x}_i-\bm {\bar{x}})^{T}}}
\end{equation}
The step-by-step implementation of physics-informed WNO is given in Algorithm \ref{alg: Training SP-PINN}.

\begin{algorithm}[ht!]
    \caption{Implementation of stochastic projection based method for physics informed WNO}\label{alg: Training SP-PINN}
    \textbf{Requirements:} Boundary conditions, initial conditions, and PDE describing the physics constrain
    \\
    {\textbf{Output:} Prediction of the field variable/solution of PDE}
    \begin{algorithmic}[1]
    \State {\textbf{Initialize:} Network parameters, $\bm{w} = \{w_{i}s, b_{i}s\}$} of the WNO, $\bm{\theta}_{NN}$.
    \State Collect the output of WNO ($\mathcal U$) and corresponding coordinates of the grid points over the domain $\{x^{i}_{f}, y^{i}_{f} \} \in \Omega$.
    \State Collect the output of WNO ($\mathcal U$) and corresponding coordinates of the grid points at the boundary of the domain $\{x^{i}_{b}, y^{i}_{b} \} \in \partial \Omega$
    \State For the given resolution of the field, define the neighborhood of each grid point points in terms of distance $\Delta x$ and $\Delta y$
    \State Obtain the first-order gradients at all the collocation points using \autoref{spgf_net2}  and store the gradients
    \State From the first-order gradients, using the same formulation \autoref{spgf_net2} the second-order gradients can be achieved. 
    \State Define the PDE loss $\mathcal{L}_{PDE}$ in terms of components of the gradients
    \State Define the Boundary loss $\mathcal{L}_{BC}$, Sum the all losses to get the total loss $\mathcal{L}_{total}$
    \While {{$\mathcal{L} > \epsilon$}}
    \State {Train the network:} $\{w_{i}s, b_{i}s\}\leftarrow \{w_{i}s, b_{i}s\}-\delta \nabla_{\bm w, \bm b}{L}(\bm w, \bm b)$
    \State {epoch= epoch $+$ 1}
    \EndWhile
    \State{Return the optimum parameters for the physics informed  WNO}
    \State {Obtain Predictions/solutions}
    \end{algorithmic}
\end{algorithm}

\section{Numerical examples}\label{sec: Numerical example}
In this section, we evaluate the efficacy and robustness of the framework with the benchmark numerical examples relevant to the various engineering systems. A relative Mean Square Error (MSE) of the test case instances is used to evaluate  the performance of the framework. In addition, we closely examine and carry out a comparative study of three distinct training scenarios of WNO; firstly, learning solely based on physics constraints; secondly, data-driven setting; and lastly, a hybrid approach in which the WNO utilizes both data and physics. In regards to the architecture of the WNO framework, it is composed of a number of layers that varies from 3–5 in accordance with the example. The ADAM optimizer with an initial learning rate of $0.001$ and a weight decay of $10^{-6}$ is used to optimize the WNO's parameters. A learning rate decays chosen to be 0.75 for every 50 epochs during training. While the total number of epochs for the example varies from 300 to 400, the batch size varies from 10 to 25. The results of considered numerical examples are summarized in the 

\begin{table}[ht!]
    \centering
    \caption{\textbf{Results of the prediction error of the physics informed WNO for all the illustrated examples}. The prediction error corresponds to the relative Mean Square Error (MSE) of the optimized model after training, computed over the test data. $N_s$ represents the number of training samples.}
    \label{Error in prediction}
    \begin{tabular}{lcccccc} 
        \toprule
        {}&\multirow{2}{*}{\textbf{PDEs}} & \multicolumn{3}{c}{\textbf{WNO}} \\ \cline{3-5}
        {}&{}& \textbf{PIWNO} & \textbf{Data-driven} & \textbf{Data + physics}  \\ \midrule
        \textbf{Relative MSE} & \multirow{2}{*}{Burger's} & $0.0518 \pm 0.041 \%$ & $0.1552 \pm 0.17 \%$ & $0.0362 \pm 0.028 \%$ \\
        $\mathbf{N_s}$ & {} & {--} & {300} & {300}\\
        \hdashline
        \textbf{Relative MSE} & \multirow{2}{*}{Nagumo} & $0.0778 \pm 0.088 \% $ & $0.0254 \pm 0.023 \%$ & $0.0202 \pm 0.020 \%$\\
        $\mathbf{N_s}$ & {} & {--} & {800} & {800}\\
        \hdashline
        \textbf{Relative MSE} & {Non-homogeneous} & $0.0469 \pm 0.015 \% $ & $ 0.0549 \pm 0.020 \%$& $ 0.0371 \pm 0.020$\\
        $\mathbf{N_s}$ & {Poisson’s} & {--} &{500} & {500}\\
        \hdashline
        \textbf{Reltive MSE} & \multirow{2}{*}{Allen-Cahn} & $2.73 \pm 0.98 \% $ & $ 2.98 \pm 0.83 \%$  &	$ 2.37 \pm 0.70$ \\
        $\mathbf{N_s}$& {} & {--} &{600} & {600}\\
        \bottomrule  
    \end{tabular}
\end{table}

\subsection{Burgers' diffusion dynamics}
As a first example, we consider the popular Burgers' equation with varying initial conditions. The 1D Burgers equation is commonly used to mathematically model the physics of weave formation, turbulence, fluid flows in fluid mechanics and gas dynamics, traffic flow and etc. \cite{kutluay1999numerical,wazwaz2002partial}. The one-dimensional (1D) Burgers equation with periodic boundary conditions is given by the following mathematical form:
\begin{equation}
    \begin{aligned}
    \partial_t u(x, t)+\frac{1}{2} \partial_x u^2(x, t) & =\nu \partial_{x x} u(x, t), & & x \in(0,1), t \in(0,1] \\
    u(x=0, t) & =u(x=1, t) =0, & & x \in(0,1), t \in(0,1] \\
    u(x, 0) & =u_0(x), & & x \in(0,1).
    \end{aligned}
\end{equation}
where $\nu = 0.1$ is the viscosity of the flow, $u_0(x) = cos(\zeta \pi x) + sin(\eta \pi x)$ is the initial condition. For generating multiple initial conditions, the parameter $\zeta$ and $\eta$ are simulated from uniform distributions with $\zeta \sim \operatorname{Unif}(0.5,1.5)$ and $\eta \sim \operatorname{Unif}(0.5,1.5)$. For different initial conditions, the goal here is to learn the operator, $\mathcal{D}: u_0(x) \mapsto u(x, t)$, where $u(x,t)$ is the Spatio-temporal solution at some time $t$. The ground truth data sets are generated employing a Matlab PDE solver, where the viscosity parameter is taken as $v=0.1$. The resolution of the solution space is chosen to be $81\times 81$. Upon training with randomly generated instances of the initial conditions, the model accuracy is evaluated with unseen realization, while a validation set of realizations are used to obtain the optimum tunable hyperparameters of the model. 
\begin{figure}[!ht]
    \centering{
    \includegraphics[width=1\textwidth]{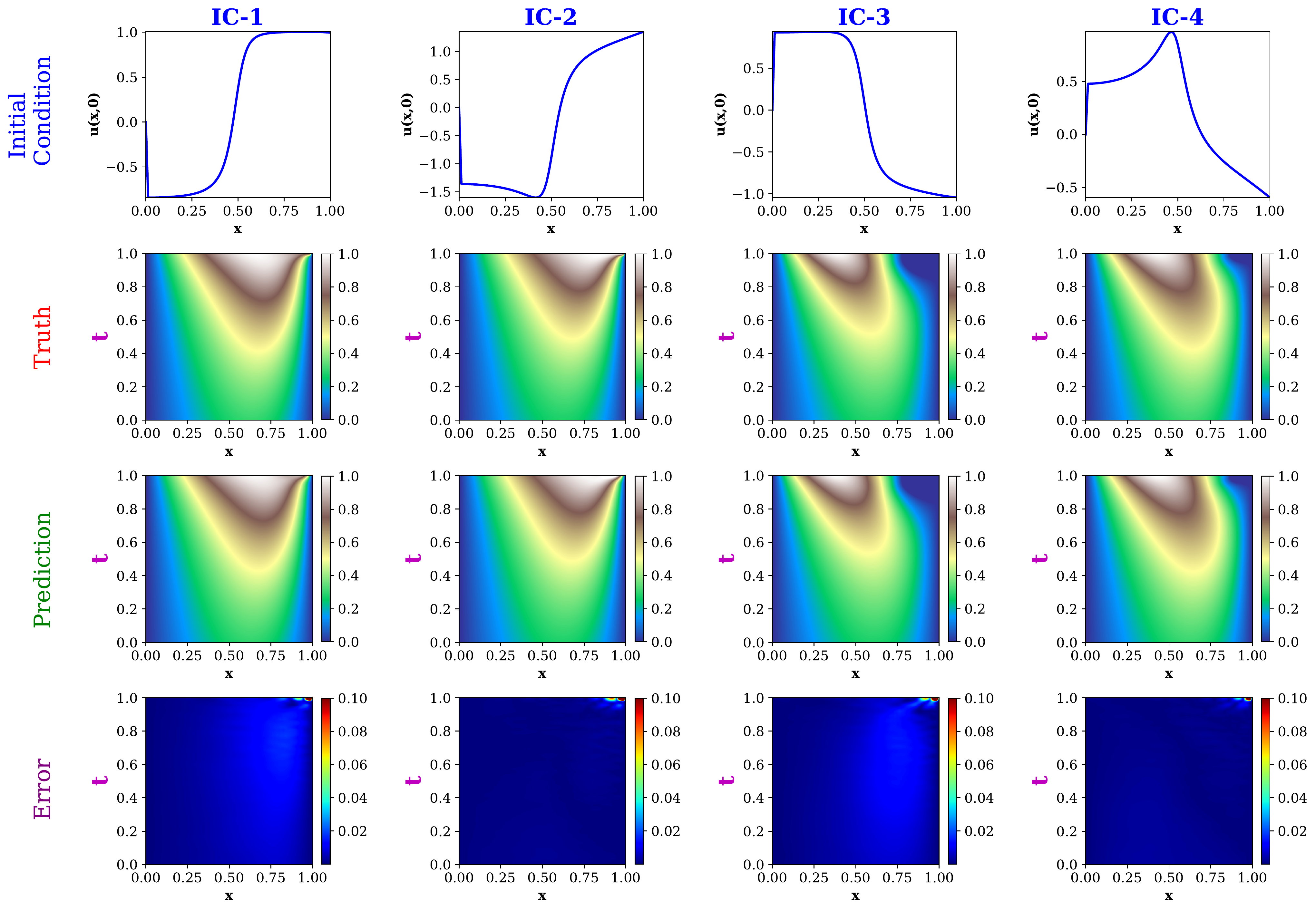}}
    \caption{The results for the parametric Burger's equation comprised of given initial conditions, corresponding ground  truth solutions, predictions, and error plots illustrated with 4 different unseen sample instances. The PIWNO maps the initial conditions to the corresponding solutions $u(x,t)$ over the domain with a Spatio-temporal resolution of $81\times81$.}\label{fig:burger}
\end{figure}

The visualization of the predicted results is showcased in the \autoref{fig:burger}, where figures comprise the input, ground truth solution, prediction, and the prediction error with reference to the ground truth over the domain for the 4 different test cases. It can be apparently observed that the predictions of PIWNO yield a remarkable concordance with the corresponding reference solutions. Moreover, the results of prediction accuracy are presented in the \autoref{Error in prediction}, where we compare the models of WNO in the three distinct training scenarios. While in the first case, in which the WNO is trained with physics constraints, the model yields an average relative MSE of $\approx 0.0518\%$, the data-driven WNO yields a mean relative MSE of $\approx 0.1552\%$. On the other hand, in the third case, when the models trained with both data loss and physics loss yield an average prediction error of $\approx 0.0362\%$, observed to be more accurate in comparison with WNO, which is either only physics informed or only data-driven.

\subsection{Nagumo equation}
\begin{figure}[!h]
    \centering{
    \includegraphics[width=1\textwidth]{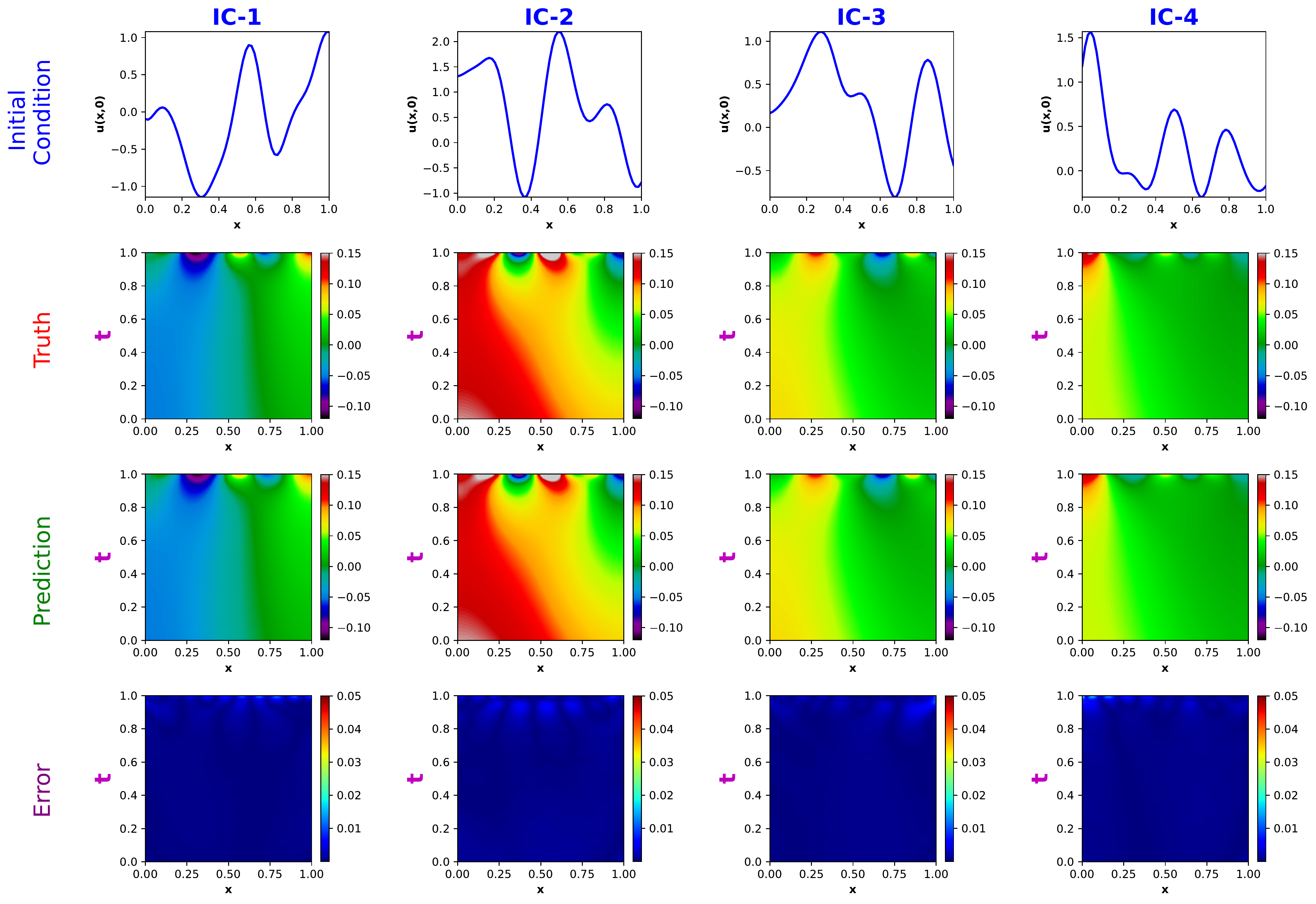}}
    \caption{The results for the parametric Nagumo equation comprised of given initial conditions, corresponding ground  truth solutions, predictions, and error plots illustrated with 4 different unseen sample instances. The physics-informed WNO maps the  initial condition to the corresponding solution $u(x,t)$ over the domain with a Spatio-temporal resolution of $65\times65$.}
    \label{fig:nagumophy}
\end{figure}

The second numerical example we consider here is the Nagumo equation. The application of the Nagumo equation can be found in modeling wave propagation in a neuron, in particular, studies involving governing dynamics of voltage across the nerve cell and the impulses in the nerve fiber. The Nagumo equation \cite{laing2009stochastic,lord2014introduction} with periodic homogeneous Neumann boundary conditions can be described as:
\begin{equation}
    \begin{aligned}
    \partial_t u - \varepsilon \partial_{x x} u &= u(1-u)(u-\alpha), & & x \in(0,1), t \in(0,1] \\
    u(x=0, t) &=u(x=1, t)=0, & & x \in(0,1), t \in(0,1] \\
    u(x, 0) & =u_0(x), & & x \in(0,1) .
    \end{aligned}
\end{equation}
where the model parameter $\alpha \in \mathbb{R}$ determines the speed of a wave traveling down the length of the axon and $\varepsilon > 0$ controls the rate of diﬀusion.
For our study, the variables are chosen to be $\varepsilon=1, \alpha=-1/2$. The  initial conditions (vector of $\mathbb{R}^{65}$) are generated using a Gaussian random field with a kernel of the form:
\begin{equation}
\mathcal{K}(x, y)=\sigma^{2} {exp}\left(\frac{{-}(\bm x-\bm{x}^{'})^{2}}{2{l}^{2}}\right)
\end{equation}
where the value of $\sigma$ and the $l$ parameters are chosen such that $\sigma= 0.1;$  and $ l = 0.1$. Moreover, to obtain the ground truth data, the PDE is numerically solved using a semi-implicit Euler method in time and centered finite differences in space to obtain the ground truth solutions. The resolution of solution space is fixed to the grid of $65\times65$. Similar to the previous example, we seek an operator that maps the initial condition $u_0(x)$ to a spatiotemporal solution at some time $t$, i.e., $\mathcal{D}: u_0(x) \mapsto u(x, t)$. To generate the instances of varying initial conditions, we simulated $u_0(x)$ from a Gaussian Random Field (GRF) \cite{bishop2006pattern}. Thereafter we used a finite difference-based script is utilized to obtain the ground truth. A detailed description of input sample generation and ground truth solver can be found in the Supplementary Materials. To this end, the performance of the proposed method is validated with the test samples of initial conditions. The results are illustrated in \autoref{fig:nagumophy}, whereas the comparative study is demonstrated in the \autoref{Error in prediction}.

The \autoref{fig:nagumophy} envisions the predicted solution and the ground truth solution. The visual resemblance of the ground truth and predicted output and the prediction error plots over the domain corresponding to the 4 unseen realizations signify the excellent performance physics informed WNO. The results of physics-informed WNO, data-drive WNO, and hybrid data and physics informed are listed in the \autoref{Error in prediction}. The observations from the table reinforce the capability of the physics WNO in learning functional mapping. However, contrary to the previous examples, the average relative MSE of the physics-informed WNO ($0.0778\%$) is more than that of the data-drive WNO($0.0254\%$). WNO trained with both data loss and physics loss on the other hand results in better model generalization in comparison with the data-driven WNO with an average MSE of ($0.0202\%$).

\subsection{Non-homogeneous Poisson's equation}
Non-homogeneous Poisson's equation is an elliptic PDE that has broad utilities in various physical situations and is considered to be the underlying mathematical model for problems, including steady-state heat diffusion and the eclectic field caused by a given electric charge. The Poisson's equation with a source term $f(x,y)$ and periodic boundary condition is expressed as:
\begin{equation}\label{Poissons_1}
 \begin{aligned}
      \partial_{xx}u + \partial_{yy}u &= f(x, y), \quad x \in[-1,1], \quad y \in[-1,1]\\
      u(x=-1, y) &=u(x=1, y)=u(x,y=-1)=u(x, y=1)=0
 \end{aligned}
\end{equation}
Here the problem under consideration is a static problem. In this example we aim to learn the operator, $\mathcal{D}: f(x,y) \mapsto u(x, y)$, where $u(x,y)$ is the solution of the equation over the $2-D$ domain. In other words here we attempt to learn operator mapping between the source function $f(x,y)$ to the solution $u(x,y)$.
An analytical solution of the form $u(x,y) = \alpha \sin(\pi{x})(1+\cos(\pi y))+ \beta \sin(2\pi x)(1-\cos(2\pi y))$ is conveniently chosen to obtain the ground truth solution. As to generate input data,  an analytical form of the source function is used such that $f(x, y) =16 \beta \pi^2\ (\cos(4\pi y)\sin(4\pi x) + \sin(4\pi x)(\cos(4\pi y) - 1))
- \alpha \pi^2(\cos(\pi y)\sin(\pi x) + \sin(\pi x)(\cos(\pi y) + 1))$. Here the expression of $f(x,y)$ is obtained by substituting the analytical expression of  $u(x,y) = \alpha \sin(\pi{x})(1+\cos(\pi y))+ \beta \sin(2\pi x)(1-\cos(2\pi y))$ in the \autoref{Poissons_1}. Thus, the boundary conditions are implicitly satisfied the solution here. The samples of the training instances are yielded by varying the variable $\alpha$ and $\beta$ such that $\alpha \sim \operatorname{Unif}(-2,2), \eta \sim \operatorname{Unif}(-2,2)$.
Further details regarding the generation of different source field instances and the corresponding ground truth solution are provided in the supplementary materials. Upon training with randomly generated source fields, the model accuracy is evaluated with unseen realization. 
\begin{figure}[!h]
    \centering{
    \includegraphics[width=1\textwidth]{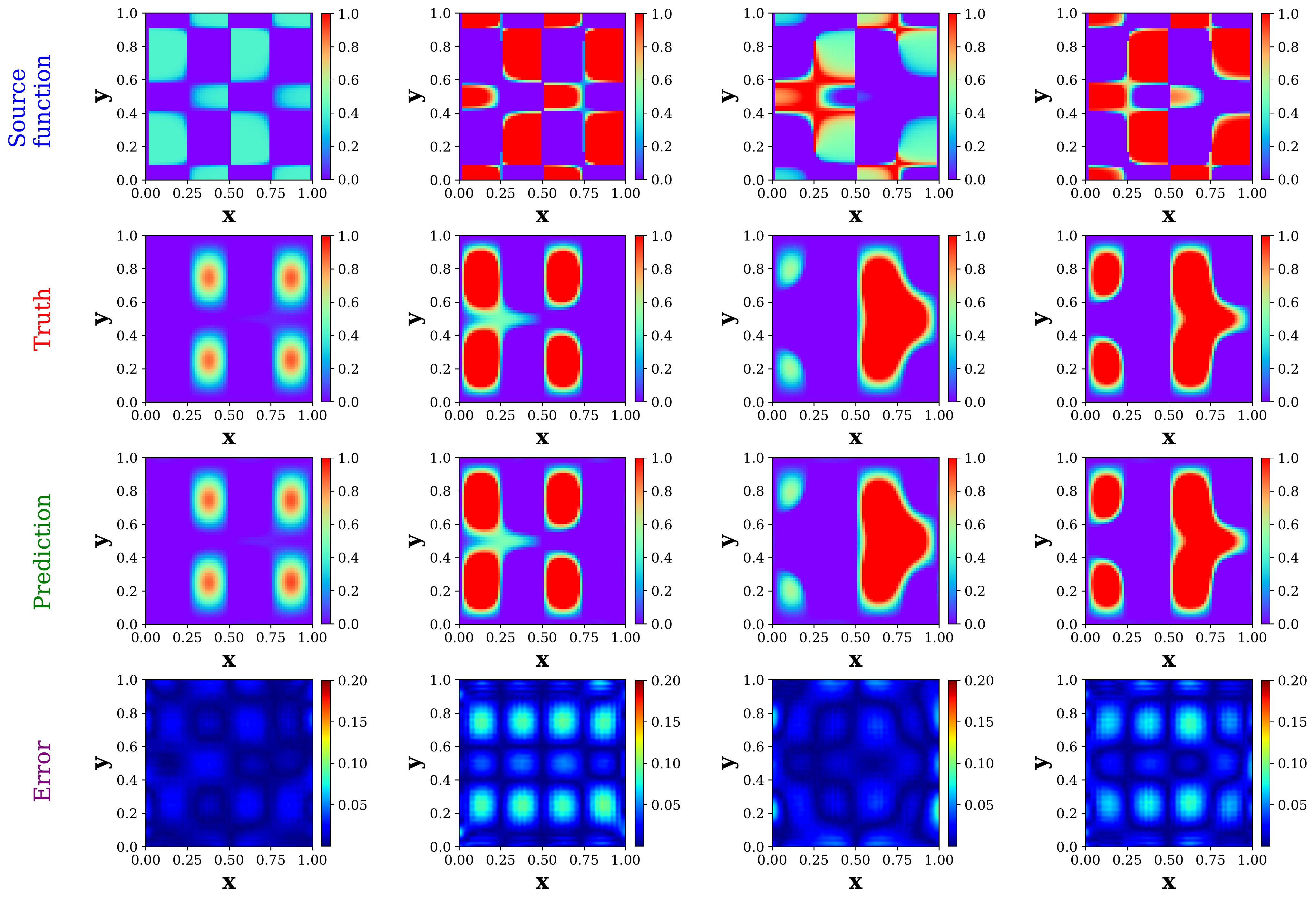}}
    \caption{The results for the Non-homogeneous Poisson's equation comprised of given source functions, corresponding ground truth solutions, predictions, and error plots illustrated with 4 different unseen sample instances. The physics-informed WNO maps the source term $f(x,y)$ to the corresponding solution $u(x,y)$ over the domain with a spatial resolution of $65\times65$.}
    \label{fig:poissons}
\end{figure}
The demonstrated results in the \autoref{fig:poissons} include the input function, ground truth solution, prediction, and prediction error. The results clearly indicate the efficacy of the proposed framework in learning the underlying PDE operator as predictions of the proposed framework in all 4 different test cases show an excellent agreement with the corresponding reference solutions. Furthermore, from the tabulated results shown in the \autoref{Error in prediction} physics-informed WNO yields an average relative MSE of $0.0469\%$, while the data-driven WNO yields a higher value of average relative MSE ($0.548\%$). In this example as well, the hybrid training using data-driven and physics-informed achieves the highest accuracy.

\subsection{Allen-Cahn equation}
As the last example, we consider the Allen–Cahn equation, which is a well-studied partial differential equation describing the phenomenon of reaction-diffusion \cite{lord2014introduction,ma2017numerical}. Applications of the mathematical model are not limited but can also be found in the context of phase separation in the multi-component alloy, chemical reactions, and crystal growth. The 2-D Allen- Cahn equation is expressed as:
\begin{equation}
\begin{aligned}
\partial_t u(x, y, t) & =\epsilon \Delta u(x, y, t)+u(x, y, t)-u(x, y, t)^3, & x, y \in(0,1) \\
u(x, y, 0) & =u_0(x, y) & x, y \in(0,1)
\end{aligned}
\end{equation}
where $\epsilon \in \mathbb{R}^{+*}$ is a real positive viscosity coefficient. For the current study, $\epsilon$ is chosen to be $\epsilon=1 \times 10^{-3}$. The problem is defined on a periodic boundary. The Gaussian Random Field with the following kernel is utilized to generate the initial conditions:
\begin{equation}
    \mathcal{K}(x, y)=\tau^{(\alpha-1)}\left(\pi^2\left(x^2+y^2\right)+\tau^2\right)^{\frac{\pi}{2}}.
\end{equation}
Here the parameters for the kernel are chosen as $\tau=15$ and $\alpha=1$. The spectral Galerkin method is used to generate the training and testing data \cite{lord2014introduction}.
In this example, we aim to learn the solution operator $\mathcal{D}: \left.\left.\bm u\right|_{(0,1)^2 \times[0,10]} \mapsto \bm u \right|_{(0,1)^2 \times(10, T]}$.
More precisely, the operator $\mathcal{D}$ here maps the temporal fields, $u(x, y, t)$, at the time steps $t \in [0, 10]$ to the solution fields, $u(x, y, t)$, at a target time step $t \in (10, T]$. For the illustration, we predict the solution till the next 10 time steps, i.e., $T=20 \mathrm{~s}$.
\begin{figure}[!h]
    \centering{
    \includegraphics[width=1\textwidth]{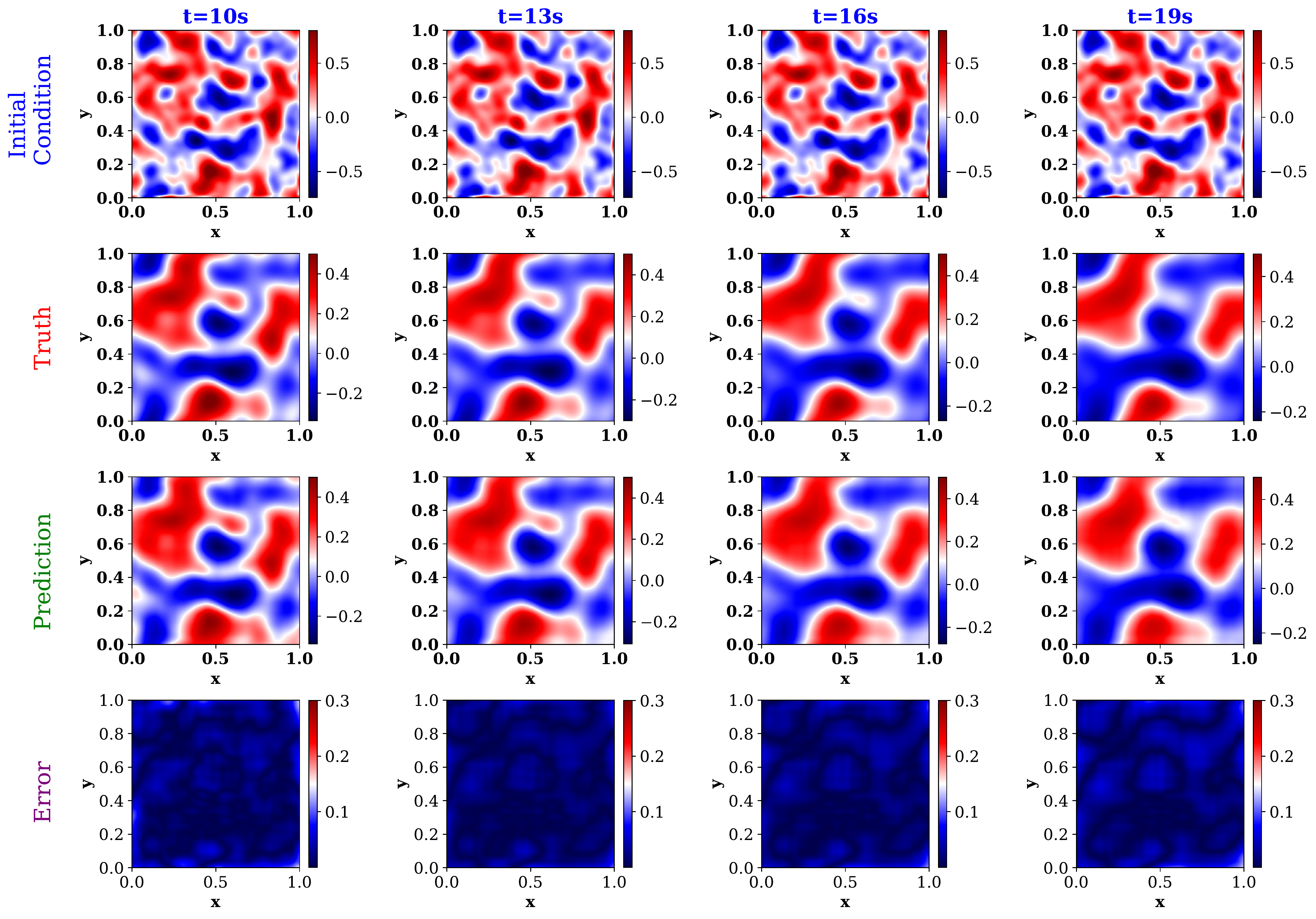}}
    \caption{The results for Allen-Cahn's equation comprised of given initial field and corresponding ground truth solutions, predictions, and error plots illustrated with an unseen sample instance at time steps $10~s,13~s,16~s$ and $19~s$. The physics-informed WNO receives the spatial field, $u(x,y)$ with a resolution of $65\times65$, for the initial 10-time steps and which maps to the corresponding solution $u(x,y)$ for the 10 future time-steps}
    \label{fig:Allencahn}
\end{figure}
A visualization of the results is provided in \autoref{fig:Allencahn}, where the predicted solution and the ground truth at times steps $t = 10 s,t = 13s, t = 16s$ and $t=19 s$ are compared. For the given unseen scenario, the prediction of the physics informed 
matches closely with the ground truth solution. Furthermore, it can be observed from the prediction error listed in the table that the average relative MSE of the physics-informed WNO is around $2.73\%$, which in comparison with previous examples, is a larger value. However, for this example, the data-driven WNO yields an average relative MSE of $(2.98\%)$, which is slightly a higher value than that of the physics-informed WNO. As seen in all previous examples, here as well, the WNO trained with both data loss and physics loss leads to greater model generalization with an average relative MSE of ($2.37\%$).

\section{Conclusions}\label{sec:Conclusions}
In this paper, we propose physics-informed WNO, a novel operator framework for approximating nonlinear operators which enables learning directly from the governing physics. Theoretically, to obtain the functional mapping in infinite dimensional Banach spaces, WNO performs a series of updates on input followed by local transformations. The operations are facilitated by the integral kernel operator in combination with nonlinear activation functions, while the parameterized integral kernel is devised as a convolutional neural network. Further, the governing physics is conveniently incorporated by employing stochastic projection-based gradients. In essence, the proposed method is capable of learning the solutions even when the problem is nonlinear and complex, which requires no corpus pair of input-output data. 

WNO possesses unified  wavelet integral layers that are able to track the finest features of signals through spatial and frequency localization. This allows the WNO to outperform the state of art operator frameworks in learning the highly nonlinear operators even when the solution domain has complex geometry and boundary conditions. While the vanilla WNO demands the output or the PDE solution corresponding to the input functions, the physics-informed WNO does not rely on the conventional PDE solvers for training. Moreover, the data-driven WNO can approximate the operator accurately only in the regimes of training data, whereas physics-informed  WNO is able to obtain a more generalized model. It is noteworthy that the proposed approach is highly efficient and is ideally suited for PDEs governing  real-time scenarios. The scope of the work is not limited but can be extended to interesting directions, such as applications of WNO for the development of digital twins and WNO-based physics augmented learning. Despite the promising performance of the WNO, it is liable to some significant drawbacks. Firstly, the absence of a standardized procedure to obtain the optimum and desired network architecture of a physics-informed WNO and to assign the weights of boundary/initial condition loss. Secondly, the lack of information regarding how to learn the operator for systems involving multi-scale physics systems. To this end, we argue that  the aforementioned challenges can be considered as the prospective goals, and addressing the limitations could result in a robust framework for the  physical systems across a wide range of science and engineering applications.
\section*{Acknowledgements}
NN and TT acknowledge the support received from Ministry of Education in the form of Prime Ministers Research Fellowship. SC acknowledges the financial support of Science and Engineering Research Board (SERB) via grant no. SRG/2021/000467.


\begin{thebibliography}{10}

\bibitem{debnath2005nonlinear}
L.~Debnath, L.~Debnath, Nonlinear partial differential equations for scientists
  and engineers, Springer, 2005.

\bibitem{jones2009differential}
D.~S. Jones, M.~Plank, B.~D. Sleeman, Differential equations and mathematical
  biology, Chapman and Hall/CRC, 2009.

\bibitem{evans2010partial}
L.~C. Evans, Partial differential equations, Vol.~19, American Mathematical
  Soc., 2010.

\bibitem{kang1996finite}
F.~Kang, S.~Zhong-Ci, F.~Kang, S.~Zhong-Ci, Finite element methods,
  Mathematical Theory of Elastic Structures (1996) 289--385.

\bibitem{cottrell2009isogeometric}
J.~A. Cottrell, T.~J. Hughes, Y.~Bazilevs, Isogeometric analysis: toward
  integration of CAD and FEA, John Wiley \& Sons, 2009.

\bibitem{ozicsik2017finite}
M.~N. {\"O}zi{\c{s}}ik, H.~R. Orlande, M.~J. Colaco, R.~M. Cotta, Finite
  difference methods in heat transfer, CRC press, 2017.

\bibitem{eymard2000finite}
R.~Eymard, T.~Gallou{\"e}t, R.~Herbin, Finite volume methods, Handbook of
  numerical analysis 7 (2000) 713--1018.

\bibitem{sirignano2018dgm}
J.~Sirignano, K.~Spiliopoulos, Dgm: A deep learning algorithm for solving
  partial differential equations, Journal of computational physics 375 (2018)
  1339--1364.

\bibitem{chan2019machine}
Q.~Chan-Wai-Nam, J.~Mikael, X.~Warin, Machine learning for semi linear pdes,
  Journal of scientific computing 79~(3) (2019) 1667--1712.

\bibitem{samaniego2020energy}
E.~Samaniego, C.~Anitescu, S.~Goswami, V.~M. Nguyen-Thanh, H.~Guo, K.~Hamdia,
  X.~Zhuang, T.~Rabczuk, An energy approach to the solution of partial
  differential equations in computational mechanics via machine learning:
  Concepts, implementation and applications, Computer Methods in Applied
  Mechanics and Engineering 362 (2020) 112790.

\bibitem{psichogios1992hybrid}
D.~C. Psichogios, L.~H. Ungar, A hybrid neural network-first principles
  approach to process modeling, AIChE Journal 38~(10) (1992) 1499--1511.

\bibitem{lagaris1998artificial}
I.~E. Lagaris, A.~Likas, D.~I. Fotiadis, Artificial neural networks for solving
  ordinary and partial differential equations, IEEE transactions on neural
  networks 9~(5) (1998) 987--1000.

\bibitem{sun2020surrogate}
L.~Sun, H.~Gao, S.~Pan, J.-X. Wang, Surrogate modeling for fluid flows based on
  physics-constrained deep learning without simulation data, Computer Methods
  in Applied Mechanics and Engineering 361 (2020) 112732.

\bibitem{zhu2019physics}
Y.~Zhu, N.~Zabaras, P.-S. Koutsourelakis, P.~Perdikaris, Physics-constrained
  deep learning for high-dimensional surrogate modeling and uncertainty
  quantification without labeled data, Journal of Computational Physics 394
  (2019) 56--81.

\bibitem{lu2019deeponet}
L.~Lu, P.~Jin, G.~E. Karniadakis, Deeponet: Learning nonlinear operators for
  identifying differential equations based on the universal approximation
  theorem of operators, arXiv preprint arXiv:1910.03193 (2019).

\bibitem{wu2020data}
K.~Wu, D.~Xiu, Data-driven deep learning of partial differential equations in
  modal space, Journal of Computational Physics 408 (2020) 109307.

\bibitem{raissi2019physics}
M.~Raissi, P.~Perdikaris, G.~E. Karniadakis, Physics-informed neural networks:
  A deep learning framework for solving forward and inverse problems involving
  nonlinear partial differential equations, Journal of Computational physics
  378 (2019) 686--707.

\bibitem{cai2021physics}
S.~Cai, Z.~Mao, Z.~Wang, M.~Yin, G.~E. Karniadakis, Physics-informed neural
  networks (pinns) for fluid mechanics: A review, Acta Mechanica Sinica 37~(12)
  (2021) 1727--1738.

\bibitem{chakraborty2021transfer}
S.~Chakraborty, Transfer learning based multi-fidelity physics informed deep
  neural network, Journal of Computational Physics 426 (2021) 109942.

\bibitem{hornik1989multilayer}
K.~Hornik, M.~Stinchcombe, H.~White, Multilayer feedforward networks are
  universal approximators, Neural networks 2~(5) (1989) 359--366.

\bibitem{margossian2019review}
C.~C. Margossian, A review of automatic differentiation and its efficient
  implementation, Wiley interdisciplinary reviews: data mining and knowledge
  discovery 9~(4) (2019) e1305.

\bibitem{goswami2020transfer}
S.~Goswami, C.~Anitescu, S.~Chakraborty, T.~Rabczuk, Transfer learning enhanced
  physics informed neural network for phase-field modeling of fracture,
  Theoretical and Applied Fracture Mechanics 106 (2020) 102447.

\bibitem{kharazmi2021hp}
E.~Kharazmi, Z.~Zhang, G.~E. Karniadakis, hp-vpinns: Variational
  physics-informed neural networks with domain decomposition, Computer Methods
  in Applied Mechanics and Engineering 374 (2021) 113547.

\bibitem{yuan2022pinn}
L.~Yuan, Y.-Q. Ni, X.-Y. Deng, S.~Hao, A-pinn: Auxiliary physics informed
  neural networks for forward and inverse problems of nonlinear
  integro-differential equations, Journal of Computational Physics 462 (2022)
  111260.

\bibitem{li2018sequential}
X.~Li, C.~Gong, L.~Gu, W.~Gao, Z.~Jing, H.~Su, A sequential surrogate method
  for reliability analysis based on radial basis function, Structural Safety 73
  (2018) 42--53.

\bibitem{kovachki2021neural}
N.~Kovachki, Z.~Li, B.~Liu, K.~Azizzadenesheli, K.~Bhattacharya, A.~Stuart,
  A.~Anandkumar, Neural operator: Learning maps between function spaces, arXiv
  preprint arXiv:2108.08481 (2021).

\bibitem{chen1995universal}
T.~Chen, H.~Chen, Universal approximation to nonlinear operators by neural
  networks with arbitrary activation functions and its application to dynamical
  systems, IEEE Transactions on Neural Networks 6~(4) (1995) 911--917.

\bibitem{lu2021learning}
L.~Lu, P.~Jin, G.~Pang, Z.~Zhang, G.~E. Karniadakis, Learning nonlinear
  operators via deeponet based on the universal approximation theorem of
  operators, Nature Machine Intelligence 3~(3) (2021) 218--229.

\bibitem{li2020neural}
Z.~Li, N.~Kovachki, K.~Azizzadenesheli, B.~Liu, K.~Bhattacharya, A.~Stuart,
  A.~Anandkumar, Neural operator: Graph kernel network for partial differential
  equations, arXiv preprint arXiv:2003.03485 (2020).

\bibitem{li2020fourier}
Z.~Li, N.~Kovachki, K.~Azizzadenesheli, B.~Liu, K.~Bhattacharya, A.~Stuart,
  A.~Anandkumar, Fourier neural operator for parametric partial differential
  equations, arXiv preprint arXiv:2010.08895 (2020).

\bibitem{wen2022u}
G.~Wen, Z.~Li, K.~Azizzadenesheli, A.~Anandkumar, S.~M. Benson, U-fno—an
  enhanced fourier neural operator-based deep-learning model for multiphase
  flow, Advances in Water Resources 163 (2022) 104180.

\bibitem{tripura2023wavelet}
T.~Tripura, S.~Chakraborty, Wavelet neural operator for solving parametric
  partial differential equations in computational mechanics problems, Computer
  Methods in Applied Mechanics and Engineering 404 (2023) 115783.

\bibitem{thakur2022multi}
A.~Thakur, T.~Tripura, S.~Chakraborty, Multi-fidelity wavelet neural operator
  with application to uncertainty quantification, arXiv preprint
  arXiv:2208.05606 (2022).

\bibitem{zhang2017time}
D.~Zhang, X.~Han, C.~Jiang, J.~Liu, Q.~Li, Time-dependent reliability analysis
  through response surface method, Journal of Mechanical Design 139~(4) (2017)
  041404.

\bibitem{boggess2015first}
A.~Boggess, F.~J. Narcowich, A first course in wavelets with Fourier analysis,
  John Wiley \& Sons, 2015.

\bibitem{navaneeth2023stochastic}
N.~Navaneeth, S.~Chakraborty, Stochastic projection based approach for gradient
  free physics informed learning, Computer Methods in Applied Mechanics and
  Engineering 406 (2023) 115842.

\bibitem{hutson2005applications}
V.~Hutson, J.~Pym, M.~Cloud, Applications of functional analysis and operator
  theory, Elsevier, 2005.

\bibitem{daubechies1992ten}
I.~Daubechies, Ten lectures on wavelets, SIAM, 1992.

\bibitem{cotter2020uses}
F.~Cotter, Uses of complex wavelets in deep convolutional neural networks,
  Ph.D. thesis, University of Cambridge (2020).

\bibitem{selesnick2005dual}
I.~W. Selesnick, R.~G. Baraniuk, N.~C. Kingsbury, The dual-tree complex wavelet
  transform, IEEE signal processing magazine 22~(6) (2005) 123--151.

\bibitem{kutluay1999numerical}
S.~Kutluay, A.~Bahadir, A.~{\"O}zde{\c{s}}, Numerical solution of
  one-dimensional burgers equation: explicit and exact-explicit finite
  difference methods, Journal of computational and applied mathematics 103~(2)
  (1999) 251--261.

\bibitem{wazwaz2002partial}
A.-M. Wazwaz, Partial differential equations, CRC Press, 2002.

\bibitem{laing2009stochastic}
C.~Laing, G.~J. Lord, Stochastic methods in neuroscience, OUP Oxford, 2009.

\bibitem{lord2014introduction}
G.~J. Lord, C.~E. Powell, T.~Shardlow, An introduction to computational
  stochastic PDEs, Vol.~50, Cambridge University Press, 2014.

\bibitem{bishop2006pattern}
C.~M. Bishop, N.~M. Nasrabadi, Pattern recognition and machine learning,
  Vol.~4, Springer, 2006.

\bibitem{ma2017numerical}
L.~Ma, R.~Chen, X.~Yang, H.~Zhang, Numerical approximations for allen-cahn type
  phase field model of two-phase incompressible fluids with moving contact
  lines, Communications in Computational Physics 21~(3) (2017) 867--889.

\end{thebibliography}

\end{document}